\documentclass[a4paper,10pt]{article}

\usepackage[utf8]{inputenc}
\usepackage[T1]{fontenc}
\usepackage[justification=justified]{caption}
\usepackage{times}
\usepackage{mathptmx}
\usepackage{ifpdf}
\usepackage{amsmath,amsfonts,amssymb}
\usepackage{graphicx,xcolor}
\usepackage{adjustbox}
\usepackage{booktabs}
\usepackage{colortbl}
\usepackage{multirow}
\usepackage{lineno}
\usepackage{fancyhdr}
\usepackage{soul}
\usepackage{geometry}
\usepackage{authblk}
\usepackage{hyperref}

\definecolor{light_grey}{rgb}{0.6, 0.6, 0.6}
\title{A large-scale multicenter breast cancer DCE-MRI benchmark dataset with expert segmentations}

\author[1,2*]{Lidia Garrucho}
\author[1]{Kaisar Kushibar}
\author[1]{Claire-Anne Reidel}
\author[1]{Smriti Joshi}
\author[1,3,4]{Richard Osuala}
\author[2]{Apostolia Tsirikoglou}
\author[5]{Maciej Bobowicz}
\author[6]{Javier del Riego}
\author[7]{Alessandro Catanese}
\author[5]{{Katarzyna Gwoździewicz}}
\author[8]{Maria-Laura Cosaka}
\author[9]{Pasant M. Abo-Elhoda}
\author[9]{Sara W. Tantawy}
\author[9]{Shorouq S. Sakrana}
\author[9]{Norhan O. Shawky-Abdelfatah}
\author[9]{Amr Muhammad Abdo-Salem}
\author[10]{Androniki Kozana}
\author[11,12]{Eugen Divjak}
\author[11,12]{Gordana Ivanac}
\author[13]{Katerina Nikiforaki}
\author[14]{Michail E. Klontzas}
\author[15]{{Rosa García-Dosdá}}
\author[16]{Meltem Gulsun-Akpinar}
\author[17]{{Oğuz Lafcı}}
\author[18]{Ritse Mann}
\author[1]{Carlos Martín-Isla}
\author[19]{Fred Prior}
\author[20, 21]{Kostas Marias}
\author[22,23]{Martijn P.A. Starmans}
\author[2,24]{Fredrik Strand}
\author[1]{Oliver Díaz}
\author[1]{Laura Igual}
\author[1, 25]{Karim Lekadir}

\affil[1]{Barcelona Artificial Intelligence in Medicine Lab (BCN-AIM), Facultat de Matemàtiques i Informàtica, Universitat de Barcelona, Gran Via de les Corts Catalanes 585 (08007), Barcelona, Spain}
\affil[2]{Department of Oncology-Pathology, Karolinska Institutet, Stockholm, Sweden}
\affil[3]{Institute of Machine Learning in Biomedical Imaging, Helmholtz Center Munich, Munich, Germany}
\affil[4]{School of Computation, Information and Technology, Technical University of Munich, Munich, Germany}
\affil[5]{2nd Dept. of Radiology, Medical University of Gdansk, Gdansk, Poland}
\affil[6]{Área de Radiología Mamaria y Ginecológica (UDIAT CD), Parc Taulí Hospital Universitari, Sabadell, Spain}
\affil[7]{Unitat de Diagnòstic per la Imatge de la Mama (UDIM), Hospital Germans Trias i Pujol, Badalona, Spain}
\affil[8]{Centro Mamario Instituto Alexander Fleming, Buenos Aires, Argentina}
\affil[9]{Department of Diagnostic \& Interventional Radiology and Molecular Imaging, Faculty of Medicine, Ain Shams University, Cairo, Egypt}
\affil[10]{Department of Radiology, University Hospital of Heraklion, Stavrakia, Greece}
\affil[11]{Department of Diagnostic and Interventional Radiology, University Hospital Dubrava, Zagreb, Croatia}
\affil[12]{University of Zagreb, School of Medicine, Zagreb, Croatia}
\affil[13]{Computational BioMedicine Laboratory, Institute of Computer Science, Foundation for Research and Technology—Hellas, Heraklion, Greece}
\affil[14]{Department of Radiology, School of Medicine, University of Crete, Heraklion, Greece}
\affil[15]{Medical Imaging and Radiology, Universitary and Politechnic Hospital La Fe, Valencia, Spain}
\affil[16]{Department of Radiology, Hacettepe University Faculty of Medicine Sihhiye, Ankara, Turkey}
\affil[17]{Department of Biomedical Imaging and Image-guided Therapy, Medical University of Vienna, Vienna, Austria}
\affil[18]{Department of Radiology and Nuclear Medicine, Radboud University Medical Center, The Netherlands}
\affil[19]{University of Arkansas for Medical Sciences, Little Rock, AR, US}
\affil[20]{Department of Electrical and Computer Engineering, Hellenic Mediterranean University, Heraklion, Greece}
\affil[21]{Computational BioMedicine Laboratory, Institute of Computer Science, Foundation for Research and Technology – Hellas (FORTH), Heraklion, Greece}
\affil[22]{Department of Radiology and Nuclear Medicine, Erasmus MC Cancer Institute, University Medical Center Rotterdam, Rotterdam, The Netherlands}
\affil[23]{Department of Pathology, Erasmus MC Cancer Institute, University Medical Center Rotterdam, Rotterdam, The Netherlands}
\affil[24]{Breast Radiology, Karolinska University Hospital, Stockholm, Sweden}
\affil[25]{Institució Catalana de Recerca i Estudis Avançats (ICREA), Passeig Lluís Companys 23, Barcelona, Spain}
\affil[*]{corresponding author: Lidia Garrucho (lgarrucho@ub.edu)}

\begin{document}

\maketitle

\begin{abstract}
Artificial Intelligence (AI) research in breast cancer Magnetic Resonance Imaging (MRI) faces challenges due to limited expert-labeled segmentations. To address this, we present a multicenter dataset of 1506 pre-treatment T1-weighted dynamic contrast-enhanced MRI cases, including expert annotations of primary tumors and non-mass-enhanced regions. The dataset integrates imaging data from four collections in The Cancer Imaging Archive (TCIA), where only 163 cases with expert segmentations were initially available. To facilitate the annotation process, a deep learning model was trained to produce preliminary segmentations for the remaining cases. These were subsequently corrected and verified by 16 breast cancer experts (averaging 9 years of experience), creating a fully annotated dataset. Additionally, the dataset includes 49 harmonized clinical and demographic variables, as well as pre-trained weights for a baseline nnU-Net model trained on the annotated data. This resource addresses a critical gap in publicly available breast cancer datasets, enabling the development, validation, and benchmarking of advanced deep learning models, thus driving progress in breast cancer diagnostics, treatment response prediction, and personalized care.
\end{abstract}

\section*{Background \& Summary}

Magnetic Resonance Imaging (MRI) is recognized as a highly sensitive imaging modality for breast cancer assessment, particularly in preoperative staging and treatment response evaluation. Breast MRI, specifically T1-weighted dynamic contrast-enhanced imaging (DCE-MRI), uses contrast agents to enhance blood vessels and tissues within the breast, aiding in the localization of tumors, often identified through angiogenesis \cite{1}. The precise delineation of the tumor boundary, or tumor segmentation, enables accurate quantitative evaluation of tumor characteristics such as shape, size, and volume, which can help monitor disease progression and treatment effectiveness. In addition to its clinical value, gold-standard segmentations enable a more nuanced analysis of breast cancer characteristics and contribute to the development of AI models for improved diagnosis and prognosis. Radiomics \cite{2}, a method widely employed in machine learning applied to radiology, involves extracting numerous quantitative features from images and is highly dependent on gold-standard segmentations \cite{3}. Although Radiomics has proven effective in predicting treatment response and survival status in breast cancer research, particularly using breast DCE-MRI images \cite{4, 5, 6}, current studies in public datasets include only a small number of subjects (up to 300) due to the limited availability of expert segmentations \cite{7}.
Besides the lack of expert segmentations, open-access DCE-MRI datasets with clinical outcomes are scarce and, currently, all available collections are part of The Cancer Imaging Archive \cite{8} (TCIA). Within the selected collections, only 163 expert primary tumor segmentations from the I-SPY1/ACRIN 6657 trial \cite{9} were available in TCIA \cite{10}. The TCIA collections lack standardization in terms of folder structure, file naming, and clinical variables.
Similarly to the M\&Ms benchmark dataset for cardiac imaging \cite{11, 12} and the BRATS dataset for brain imaging \cite{13}, our initiative presents a multicenter breast cancer dataset comprising 1506 pre-treatment DCE-MRI cases with expert segmentations. This dataset is specifically designed to support the benchmarking of advanced medical imaging models leveraging AI.

\begin{figure}[htb]
    \includegraphics[width=\linewidth]{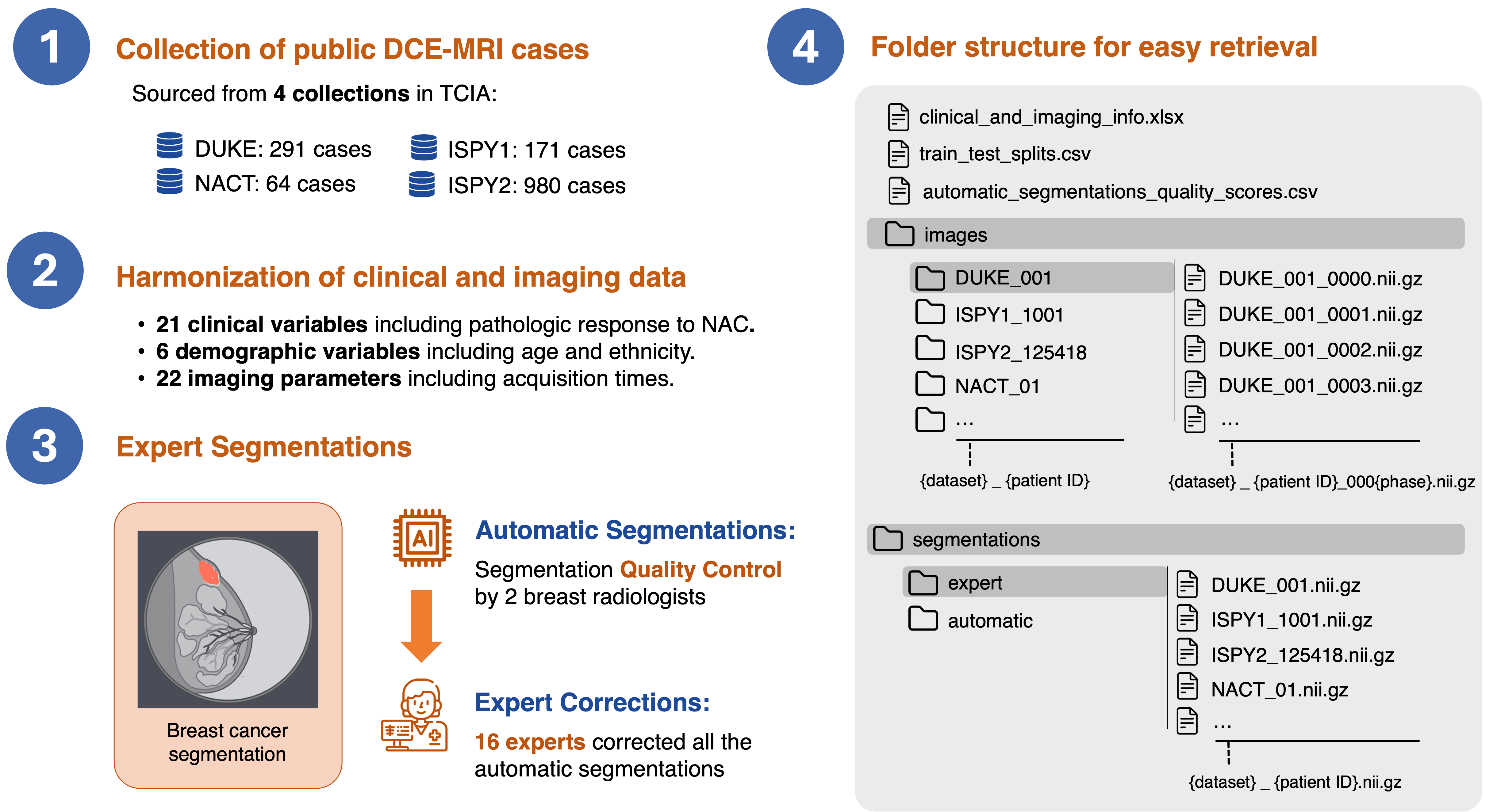}
    \centering
    \caption[]{Summary of the main contributions in the MAMA-MIA dataset. The dataset includes three tables with the harmonized clinical and imaging data, train and test splits for benchmarking, the automatic segmentation quality scores, the images and the expert and the automatic preliminary segmentations. Each case in the dataset consists of a pre-treatment T1-weighted DCE-MRI sequence with all the phases in a subfolder under the images folder, and two primary tumour segmentations, one expert-corrected and the preliminary automatic segmentation, without expert corrections.}
    \label{fig:dataset_description}
\end{figure}

The main contributions of our work are shown in Figure~\ref{fig:dataset_description}.

\begin{figure}[htb]
    \includegraphics[width=\linewidth]{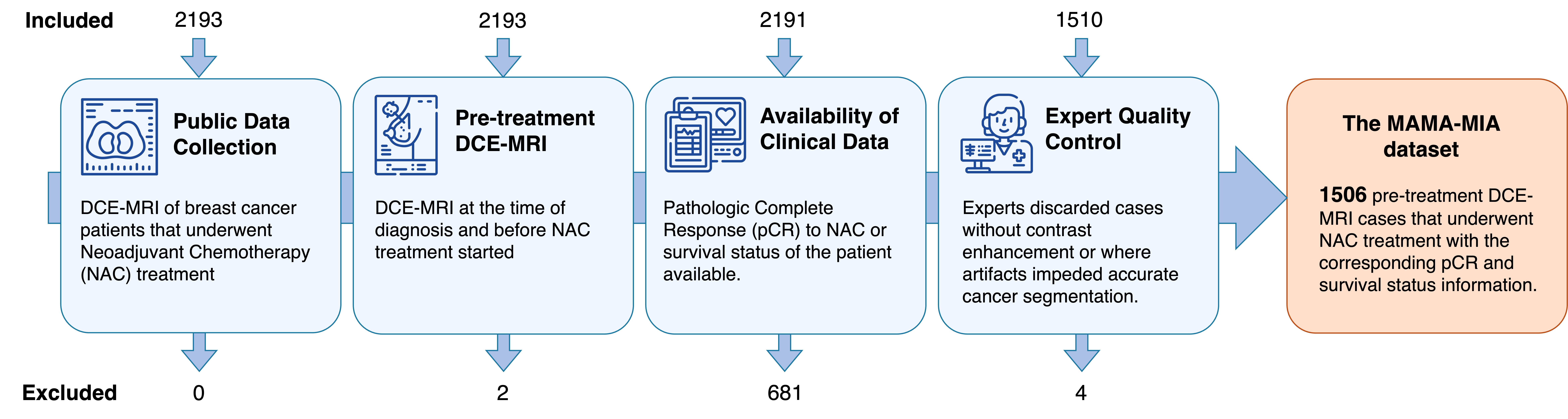}
    \centering
    \caption[]{Selection criteria to build the MAMA-MIA dataset. The DCE-MRI cases are collected from four public collections available on TCIA \cite{8}: \textit{level 2b} cohort \cite{24} from I-SPY1/ACRIN 6657 trial (I-SPY1) \cite{9}, I-SPY2/ACRIN 6698 trial \cite{25, 26}, NACT-Pilot \cite{27}, and Duke-Breast-Cancer-MRI \cite{28}, referred to as NACT, ISPY1, ISPY2, and DUKE, respectively.}
    \label{fig:selection_protocol}
\end{figure}

First, we collected pre-treatment T1-weighted DCE-MRI cases from four different collections in TCIA, sourcing a total of 1506 cases. The selection criteria, shown in Figure~\ref{fig:selection_protocol}, included pre-treatment cases where patients underwent neoadjuvant chemotherapy (NAC) within months of diagnosis and for which corresponding clinical data was available, such as the pathological complete response (pCR) to NAC or five-year survival information.

Second, the clinical and imaging data of the selected cases from four collections were consolidated and harmonized into a single table, containing 21 clinical variables, 6 demographic variables, and 22 imaging parameters.

Third, a total of 16 experts participated in the segmentation of the primary tumors and non-mass-enhanced areas present in the 1506 T1-weighted DCE-MRI cases. Manual segmentation of breast tumors in 3D MR images is both tedious and time-consuming. To facilitate this process, we initially trained a standard state-of-the-art deep learning model using private expert segmentations of DCE-MRI. This model produced preliminary segmentations that the 16 experts manually corrected, inspected, and verified, resulting in 1506 expert segmentations. Additionally, two expert clinicians visually assessed all the preliminary automatic segmentations for quality assurance. The quality labels from these automatic segmentations, along with expert corrections, offer insights for the design of AI-driven models for segmentation quality control.

Forth, the dataset folder structure was standardized for easy retrieval and harmonized to support plug-and-play AI training.

Last, an additional contribution of this work is the pre-trained weights of a baseline nnU-Net \cite{14} tumor segmentation model, trained on the 1506 expert segmentations of the primary tumors segmented in the MAMA-MIA dataset. These weights can be used for inference or to fine-tune models for a wide variety of segmentation tasks involving MRI or other 3D medical image modalities.

We note that our dataset may have potential biases arising from preliminary automatic segmentations and inter-annotator variability among the 16 experts performing manual corrections. Typically, radiologists employ similar software tools (e.g., thresholding) to generate rough approximations of lesions, saving annotation time. Despite these potential biases, our dataset represents the largest collection of expert segmentations in breast cancer MRI to date, paired with harmonized imaging and clinical data. This addresses a significant gap in the availability of gold-standard segmentations in breast MRI, adding substantial value to breast cancer research.
It is important to highlight that experts were instructed to segment only the primary lesion in cases of multifocal or multicenter breast cancers. This restriction was due to the clinical information, such as tumor subtype and pathologic complete response (pCR), being available only for the primary lesion and to reduce segmentation time.

In the following paragraphs some of the most important potential applications of the MAMA-MIA dataset are introduced in detail.
\paragraph{Treatment Response and Survival Prediction.} Despite its benefits, NAC has associated side effects, making it desirable to predict patient response before treatment planning. Most deep learning methods predicting pCR to NAC using MRI data have been developed with fewer than 300 samples and are difficult to benchmark due to the lack of a standardized dataset with expert segmentations \cite{7}. By including treatment and survival outcomes alongside other clinical variables, the MAMA-MIA dataset can serve as a benchmark for developing AI models to predict treatment response and patient survival.
\paragraph{Automatic Segmentation of Breast Cancer in MRI.}
Automated segmentation algorithms can process medical images much faster than manual methods, minimizing inter-observer variability and providing more consistent results. The 1506 expert segmentations in this dataset enable the development of large-scale, generalizable, and robust automatic tumor segmentation models. The pre-trained weights of a baseline nnU-Net segmentation model trained on this dataset are provided to facilitate further improvements and to serve as a baseline for comparative studies.
\paragraph{Segmentation Quality Control.}
Visual inspection by expert radiologists is the gold standard for quality control, but it is impractical at scale \cite{15}. The expert segmentations, combined with evaluations of automatic segmentations, form the basis for developing robust quality control mechanisms in breast cancer MRI.
\paragraph{Image Synthesis.}
The synthesis of realistic and diverse 2D MRI slices, as well as full 3D DCE-MRI volumes, can enhance the optimization of image analysis algorithms via data augmentation, domain adaptation, or privacy preservation. This can also support radiologist decision-making, such as simulating treatment response or predicting disease progression \cite{16, 17, 18, 19}. Patient demographics and clinical data in the dataset can be utilized to condition generative models or analyze their impact on generated images, enabling AI fairness analysis and bias mitigation (e.g., age or ethnicity).
\paragraph{Image Standardization.}
The dataset includes both bilateral and unilateral images, with variations in magnetic field strengths, number of slices, slice thickness, and scanner manufacturers, making it a valuable resource for developing domain generalization and image standardization techniques. Exploring contrast dynamics in tumors and their correlation with acquisition times, as included in the harmonized imaging data, represents a promising avenue for future studies.
\paragraph{Fine-tuning of Foundational Models.} Foundation models like MedSAM \cite{20}, based on SAM \cite{21}, address segmentation tasks across imaging modalities but suffer from imbalances in modality representation in training data and challenges with segmenting vessel-like structures. While SAM has been explored for breast tumor segmentation in ultrasound images \cite{22} and mammographic mass segmentation \cite{23}, the MAMA-MIA dataset could accelerate the ingestion of 3D medical imaging data for training or fine-tuning foundational models tailored to breast MRI tasks.

\section*{Methods}

\subsection*{Data Collection and Harmonization}

The steps to collect the DCE-MRI cases that form the MAMA-MIA data set are illustrated in Figure~\ref{fig:selection_protocol}. The initial selection criterion was to gather all available open-access DCE-MRI studies of breast cancer patients who underwent NAC treatment. Four collections available on TCIA \cite{8} met this requirement: the \textit{level 2b cohort} \cite{24} from the I-SPY1/ACRIN 6657 trial (I-SPY1) \cite{9}, the I-SPY2/ACRIN 6698 trial \cite{25, 26}, NACT-Pilot \cite{27}, and Duke-Breast-Cancer-MRI \cite{28}, referred to as ISPY1, ISPY2, NACT, and DUKE, respectively.

The second criterion was to select the DCE-MRI series acquired before NAC treatment started, often referred to as pre-treatment sequences or timepoint T0. The third criterion was to exclude cases that lacked information on treatment response or survival status. The final criterion involved the quality control by experts during cancer segmentation in the DCE-MRI images. Experts excluded cases without sufficient contrast enhancement or those with artifacts that significantly impeded segmentation.

\begin{figure}[htb]
    \includegraphics[width=\linewidth]{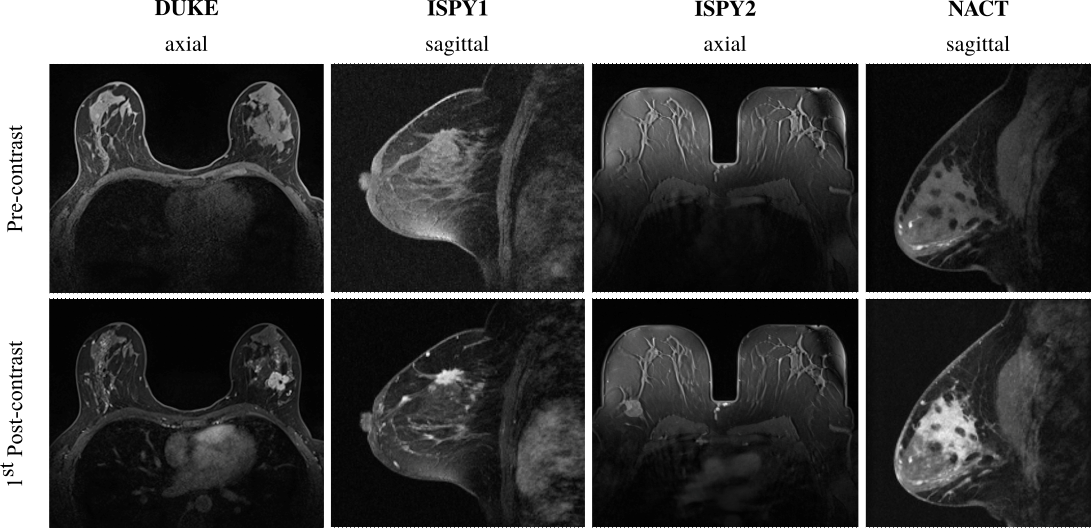}
    \centering
    \caption[]{Pre-treatment DCE-MRI sequences from the four collections forming the dataset. Left to right: images are shown in the acquisition plane (axial or sagittal) from DUKE, ISPY1, ISPY2 and NACT. Only two phases of the DCE-MRI sequence are shown, the pre-contrast phase (first row) and the first post-contrast phase (second row).}
    \label{fig:full_images}
\end{figure}

The final MAMA-MIA dataset comprises 1506 DCE-MRI cases that meet all the selection criteria. Figure~\ref{fig:full_images} illustrates the pre-contrast and post-contrast phases of one case from each of the four different collections included in the dataset. In the first post-contrast phase, malignant tissues are better visualized due to enhanced contrast after injection. The dataset includes both bilateral and unilateral breast MRIs, as well as images acquired in axial and sagittal planes. This diversity arises from scanner-dependent acquisition protocols at different centers, where some scanners capture MRIs in the axial plane, while others use sagittal protocols. By including this variety, the dataset reflects real-world clinical practices, enhances its generalizability, and supports research into a wide range of imaging scenarios.

The dataset harmonization steps included data curation, image quality control, extraction of clinical and imaging data from DICOM headers, and establishing a standardized naming and folder structure for all the sequences in the dataset. In addition, the orientation was standardized to ensure consistency and facilitate usability in computational analysis. Axial MRIs were reoriented to the LPS (left-posterior-superior) coordinate system, while sagittal MRIs were reoriented to the RAS (right-anterior-superior) coordinate system. This decision aligns with common practices in machine learning and deep learning research, where software such as nibabel\cite{29} and SimpleITK\cite{30} is frequently used to process medical imaging data. Reorienting the images in this manner ensures that the X, Y, and Z axes consistently correspond to axial, coronal, and sagittal views, respectively, regardless of the original acquisition orientation. This uniformity simplifies downstream tasks such as visualization, annotation, and model development, while preserving anatomical accuracy and clarity for each imaging plane.

No additional preprocessing steps, such as bias field correction, image normalization (e.g., z-score or min-max normalization), or voxel resampling, were applied to the dataset. These methods can significantly alter image resolution and potentially impact model performance. Preprocessing choices depend on the specific requirements of the downstream task \cite{31}. For instance, voxel resampling may influence spatial resolution crucial for tumor segmentation, while normalization strategies might affect intensity-based tasks like radiomics or contrast enhancement studies. Researchers can tailor these procedures to their use cases.

\subsection*{Expert Segmentations}\label{sec:expert_seg_section}
The dataset cohort includes a highly heterogeneous group of locally advanced breast cancers, encompassing cases with single tumors, multiple tumors (multifocal cases), non-mass enhanced areas where the cancer has spread, and bilateral breast cancers. In our dataset, both automatic and expert segmentations were performed within the Volume of Interest (VOI), excluding other cancerous findings outside the VOI. The reason for segmenting only the tissues within the VOI is related to the clinical outcomes and tumor subtype information, which is available only for the primary tumor (delineated VOI) and not for bilateral or multifocal breast cancers.

\subsubsection*{Selection of the Volume of Interest}\label{sec:ptv_section}
The VOI is defined as a 3D rectangular box manually drawn to encompass the entire enhanced region. Its dimensions depend on the tumor morphology, ranging from a few centimeters to the entire breast in cases of more advanced tumors. Accurate selection of the VOI is important for segmentation, as the clinical information regarding tumor subtype and treatment response can only be reliably linked to the volume within the VOI.

In DUKE, the bounding boxes are provided within the clinical information. However, not all datasets include straightforward 3D coordinates of the tumor. For the NACT, ISPY1, and ISPY2 collections, tumor volumetric analysis images are available in most of the DCE-MRI cases. These volumetric analysis images include various annotations of the breast tissue and pixel-level annotations of the peak enhanced region after contrast injection, also known as the Functional Tumor Volume (FTV).

\begin{figure}[htb]
    \includegraphics[width=0.7\linewidth]{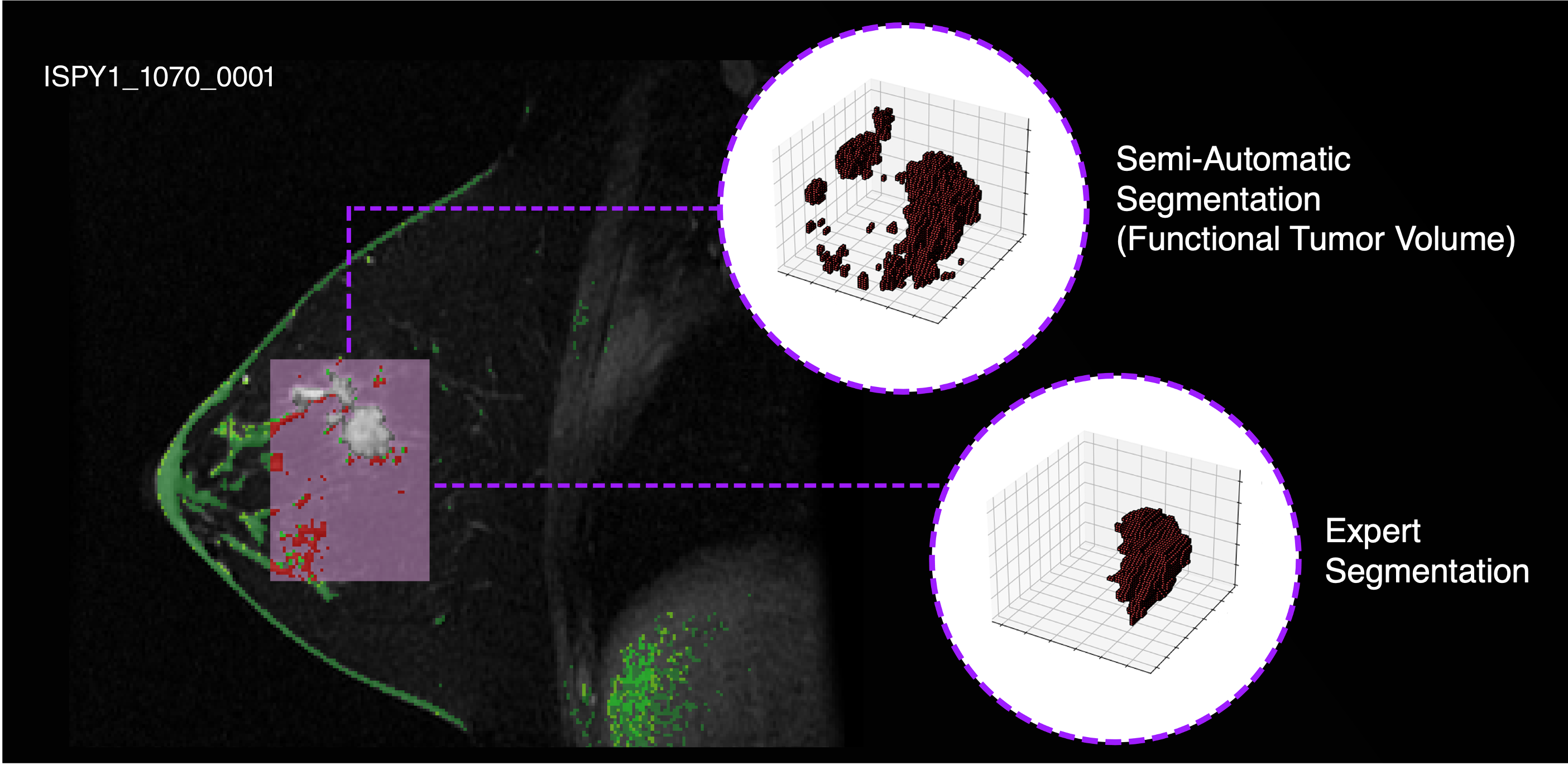}
    \centering
    \caption[]{Volumetric analysis image with the corresponding tumor bounding box (in purple) and the tumor volume extracted using the signal enhancement ratio (SER) method denoted as Functional Tumor Volume (FTV). In comparison to FTV, the expert segmentation of the tumor is more precise and contains only the malignant tissues.}
    \label{fig:ftv_mask}
\end{figure}

The FTV is determined by filtering the percent enhancement (PE) image and the signal enhancement ratio (SER) image using specific thresholds on their pixel values, following the steps described in \cite{4}. Figure~\ref{fig:ftv_mask} compares the FTV to the expert tumor segmentation. Although the FTV delineates a region likely to encompass malignant tissue, it frequently fails to accurately represent the precise tumor volume. Consequently, expert refinement of the FTV masks is essential to ensure anatomical accuracy and reliable segmentation for downstream analyses.

In cases where the analysis mask was unavailable, an approximate VOI was created using the same filtering steps applied to the SER and PE images, available in all the remaining cases. Using these procedures, we obtained 3D bounding boxes, or VOIs, encapsulating the primary tumor or non-mass-enhanced area for all 1506 cases in the dataset.

\subsubsection*{Preliminary Automatic Segmentations}

In this study, the preliminary automatic tumor segmentations were generated using the popular nnU-Net framework \cite{14}. A segmentation model was trained using a total of 331 primary tumor and non-mass enhanced (NME) segmentations from DUKE\cite{28} and the TCGA-BRCA collection \cite{32}.
The training dataset encompasses 251 axial DCE-MRI cases from DUKE with expert segmentations shared by the authors from a treatment response study\cite{30, 6} and other 80 sagittal DCE-MRI cases with expert-validated automatic tumor segmentations \cite{34} (Chicago Dynamic MRI Explorer 2005 Version) from TCGA-BRCA, increasing the heterogeneity of the training data. This combination of axial and sagittal cases in the training dataset introduced diversity in imaging orientations, making the preliminary segmentation model more robust for segmenting the MAMA-MIA dataset, which contains MRIs acquired in both orientations. At the time this model was developed, the DUKE and TCGA datasets were the only sources of expert tumor segmentations available to the research team.

The 331 expert-validated tumor segmentations were performed on the first post-contrast images. Due to negligible patient movement, these segmentations were applied across all DCE-MRI phases, including pre- and post-contrast images, serving as additional data for training.
Preprocessing steps prior to training included cropping the images to the Volume of Interest (VOI) and resampling to $1 \times 1 \times 1 \ mm^3$ isotropic pixel spacing. Data augmentation was performed by adding a 25\% pixel margin to the VOI and applying random flipping.

The preliminary automatic segmentations were up-sampled and mapped back to the original image space to generate full-size segmentation masks of the primary lesions.
The nnU-Net model achieved a mean validation Dice coefficient of $0.8287 \pm 0.0112$ in a 5-fold cross-validation setting. 
As a note, the DCE-MRI cases from TCGA-BRCA collection were not included in the final MAMA-MIA dataset due to the absence of clinical information, such as tumor subtype, treatment response, or survival status, which is crucial for the study’s objectives.

\subsubsection*{Visual Quality Control of the Preliminary Automatic Segmentations}\label{sec:expert_qc_section}
Two expert breast radiologists evaluated the quality of the preliminary automatic segmentations using an in-house graphical user interface (GUI). For each case, three 2D slices from the first post-contrast image—spanning the axial, sagittal, and coronal planes—were displayed with segmentation contours highlighted in red. Full image slices were provided to help the experts quickly identify whether the primary tumor or NME region was missed by the segmentation model.

The visual inspection process was guided by specific instructions provided to the expert breast radiologists. Based on the different images displayed, the experts assessed the 1506 preliminary automatic segmentations and categorized them into four quality categories: \textit{Good}, \textit{Acceptable}, \textit{Poor}, or \textit{Missed}. A \textit{Good} segmentation indicated precision with no need for major corrections. An \textit{Acceptable} segmentation captured the tumor but required improvement, with only a few incorrect pixels. A \textit{Poor} segmentation lacked precision and contained numerous pixels outside the tumor region. Finally, a segmentation categorized as \textit{Missed} corresponded to an area of the breast unrelated to the tumor.

This categorization was also used to evenly distribute segmentations among the experts for correction, ensuring that the difficulty of cases assigned to each radiologist was balanced. Notably, all cases, including those rated as \textit{Good} during the visual assessment, were sent to the radiologists for manual correction without disclosing the assigned quality category (\textit{Good}, \textit{Acceptable}, \textit{Poor}, or \textit{Missed}). This blinded approach ensured that no bias was introduced during the correction process, preserving the integrity of the evaluation and ensuring that expert corrections remained consistent across categories.

\subsubsection*{Expert Manual Corrections}
\begin{figure}[htb]
    \includegraphics[width=\linewidth]{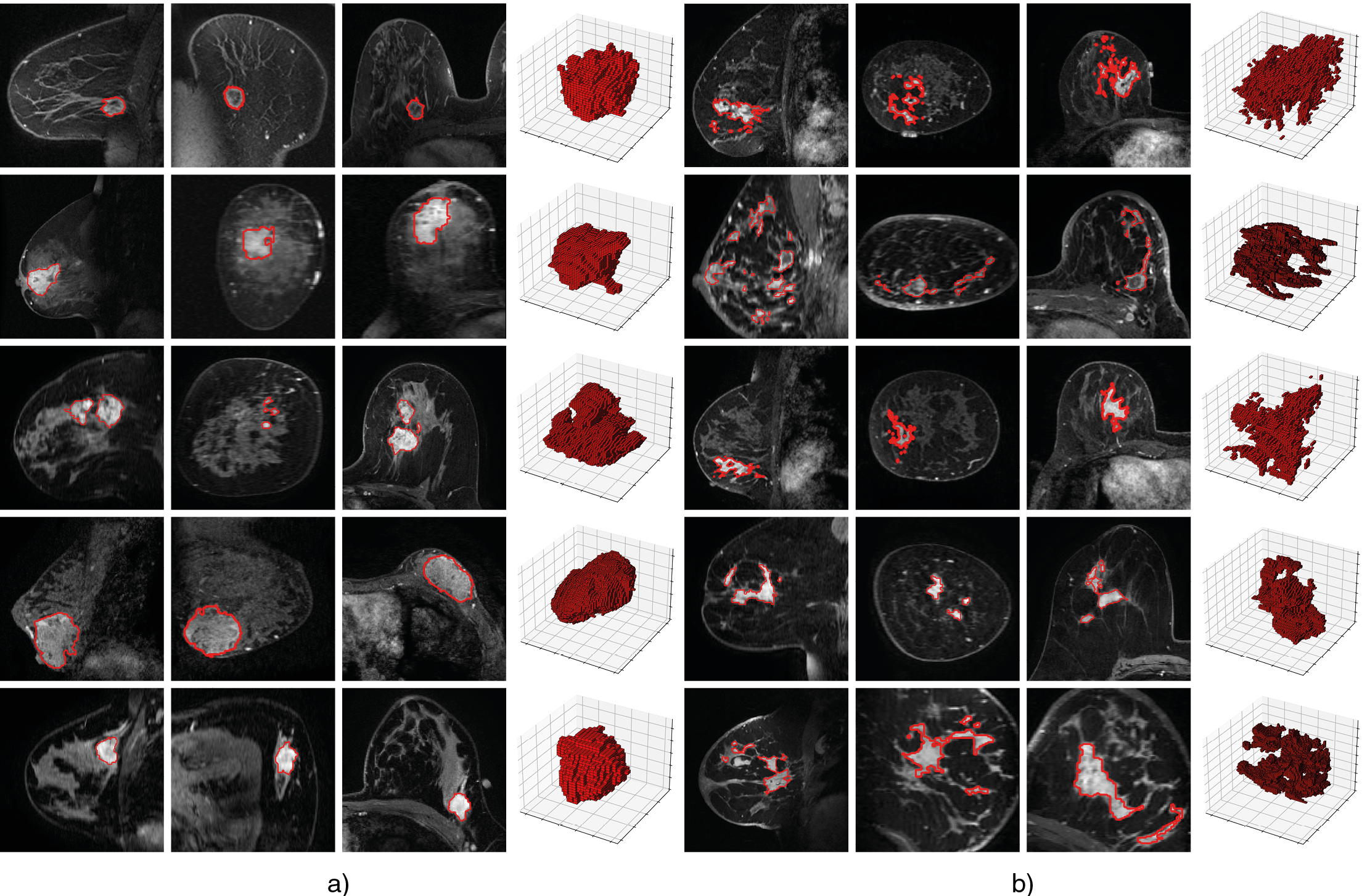}
    \centering
    \caption[]{Manual segmentations were performed for both a) primary tumors and b) non-mass enhanced areas. For each case, the middle slice of the manual segmentation is displayed in sagittal, coronal, and axial views, with the segmentation contour highlighted in red. The rightmost columns in a) and b) present the corresponding 3D segmentation. The images from the first columns correspond to different examples of primary tumor segmentations, meanwhile, the images of the second cases show the more challenging segmentation of non-mass enhanced cancers.}
    \label{fig:demo_images}
\end{figure}

From the 1506 DCE-MRI cases forming the MAMA-MIA dataset only a total of 160 manual segmentations from the ISPY1 collection \cite{10} were available in the TCIA platform. Additionally, the authors from a treatment response study using DUKE dataset \cite{30, 6} shared the expert manual segmentations from an additional 251 cases included in MAMA-MIA dataset. Therefore, a total of 411 out of 1506 cases had expert manual segmentations. Our main contribution, together with the dataset harmonization, are the manual segmentations of the missing 1095 cases.
A total of 16 experts from nine different institutions from Europe and Africa participated in the manual correction of the 3D segmentations. The group, with an average of 9 years of expertise in breast cancer radiology, was formed by fourteen breast radiologists, one surgical oncologist and one medical physicist. The preliminary automatic segmentation quality scores from one expert were used to stratify the cases and assign each expert an evenly distributed set of 70 cases in terms of corrections needed. However, as previously stated, this categorization was blind to the manual annotators. Along with the automatic segmentations, each case consisted of the pre-contrast and the first post-contrast phase. The experts were asked to segment the tumor in the first post-contrast phase but the subtracted image or a later phase could be used as a support in the manual correction process.
The Mango viewer~\cite{35} was the tool selected to correct the automatic segmentations. The guidelines provided to the experts included: 1) segment only the primary tumor if the secondary tumors are not within the FTV volume in multifocal cases, 2) avoid as much as possible the inclusion of healthy tissue in non-mass enhanced cases, 3) exclude tissue markers (or clips) from the segmentations, 4) include tumor necrosis in the segmentation, 5) do not include intra-mammary lymph nodes, 6) verify the segmentation is consistent in all views, not only in the highest resolution view.
In Figure~\ref{fig:demo_images} some examples of first post contrast images and the corresponding manual segmentations are shown.

\subsubsection*{Baseline Segmentation Model using the Expert Segmentations}

An additional contribution of this work is the pretrained weights of a vanilla nnU-Net \cite{14} tumor segmentation model, trained using the 1506 expert segmentations. The nnU-Net model was trained with DCE-MRI full-images as input over a total of 1000 epochs in a five-fold cross-validation setting. The model achieved a mean validation Dice coefficient of $0.7620 \pm 0.2113$.

The preprocessing steps included z-scoring the DCE-MRI images using the mean and standard deviation of all its phases (pre- and post-contrast) and resampling to $1 \times 1 \times 1 \ mm^3$ isotropic pixel spacing. In the training pipeline, all post-contrast phases and the subtraction MRI image (computed by subtracting the pre-contrast image from the first post-contrast image) were included as data augmentation. The model was evaluated only on the first post-contrast phases, which are the images used by the experts to perform the segmentations.

\section*{Data Records}

All data records, including the DCE-MRI images, the automatic and expert segmentations for each of the 1506 cases in the MAMA-MIA dataset and the weights of the pretrained segmentation model, are available online in the MAMA-MIA Synapse repository \cite{36} under CC-BY-NC license at the level of the most restrictive primary dataset (Duke-Breast-Cancer-MRI \cite{28}).
Data records also include three tables, one that contains all the clinical and imaging information, another with the automatic segmentation quality scores from two experts who evaluated the automatic segmentations using the GUI, and a table with the train and test split to promote reproducibility in future studies using the dataset. 
Figure~\ref{fig:dataset_description} illustrates the file content and folder structure of the MAMA-MIA dataset. Each case identifier consists of the original collection/dataset acronym and the corresponding patient identification number (patient ID). For instance, the \textit{ISPY1\_1221} case corresponds to the pre-treatment DCE-MRI sequences of patient ID 1221 from the ISPY-1 collection. The different phases are named using the same case ID plus the corresponding phase number (\textit{ISPY1\_1221\_000X}). For example, \textit{ISPY1\_1221\_0000} represents the pre-contrast phase and \textit{ISPY1\_1221\_0002} represents the second post-contrast phase.

Table~\ref{tab:social_clinical_stats} and Table~\ref{tab:imaging_stats} summarize the most representative dataset demographics, clinical variables, and image acquisition parameters. As described in Table~\ref{tab:social_clinical_stats}, age and ethnicity information is available for more than 98\% of the cases, while Body Mass Index (BMI) is available for 83\% of cases. The MAMA-MIA dataset comprises 314 cases from women younger than 40 years old, constituting 21\% of the total patients, and half of the dataset patients were younger than 50 years old at diagnosis. Therefore, the MAMA-MIA dataset can be considered well-balanced in terms of the young versus older population. Ethnicity distributions in the dataset are reflective of United States demographics~\cite{37}, with 16\% African American patients, less than 6\% Asian and other ethnicities, and a majority Caucasian population (74.9\%).

Clinical information available in more than 90\% of the cases includes the presence of bilateral cancer at diagnosis, multifocal cancer, tumor subtype, and pathological complete response (pCR) after neoadjuvant chemotherapy (NAC) treatment. Other relevant information included in the dataset, albeit not present in all cases, comprises survival status in over 450 cases, different tumor receptors, days to recurrence or metastasis, agents prescribed during NAC, the necessity of mastectomy after treatment, and more. A comprehensive list of clinical and imaging variables included in the dataset can be found in the Excel Table as part of the Supplementary Material.

\begin{figure}[htb]
    \includegraphics[width=1\linewidth]{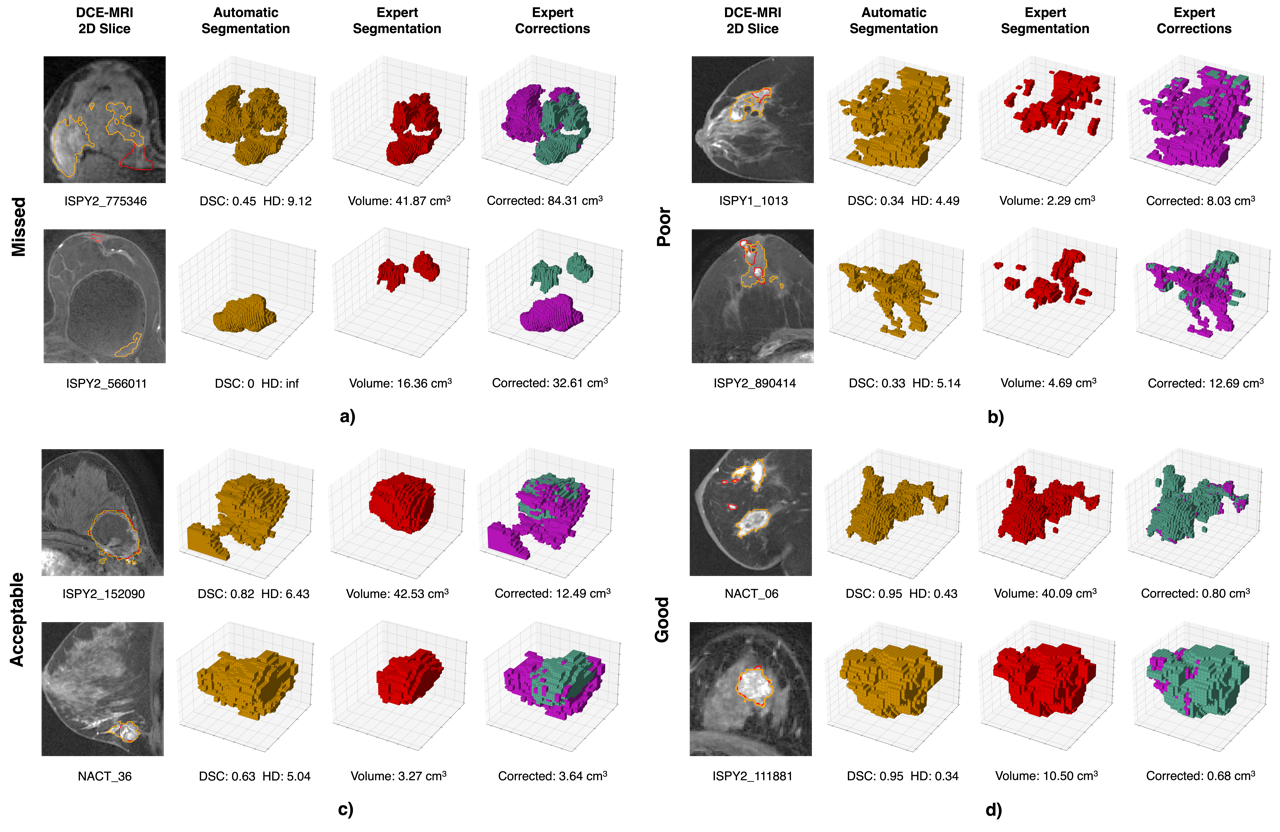}
    \centering
    \caption[]{Examples of preliminary automatic (yellow) and expert segmentations (red) and the corresponding expert corrections (pink if the voxels were removed, and green if they were added). The examples are grouped by the expert-assigned quality scores for automatic segmentations during the visual inspection (a) Missed, b) Poor, c) Acceptable, and d) Good). Under the automatic segmentation there is the Dice Similarity Coefficient (DSC) and the 95 percentile Hausdorff Distance (HD) between automatic and expert segmentations. Also, under the manual segmentation, the final volume of the expert segmentation in $cm^3$. The total volume corrected by the experts is quantified in $cm^3$.}
    \label{fig:expert_corrections}
\end{figure} 

Table~\ref{tab:imaging_stats} presents the most common imaging information included in the dataset: date of original collections, acquisition plane, magnetic field strength (Tesla) used for DCE-MRI acquisition, scanner manufacturers and models, number of bilateral and fat-suppressed DCE-MRIs, mean number of slices, slice thickness, pixel spacing, number of phases, and total number of cases obtained from each original collection. In the DCE-MRI medical imaging modality, the acquisition time interval between contrast administration (pre-contrast MRI) and subsequent post-contrast phases is an important factor. Table~\ref{tab:ph_times} summarizes the average time intervals between phases per dataset. It should be noted that older datasets like ISPY1 and NACT have longer acquisition intervals than later collections.

\section*{Technical Validation}

This technical validation evaluates the visual quality control of preliminary automatic segmentations, emphasizing the role of expert-corrected segmentations as the primary contribution of this study. The goal is to provide actionable guidelines and recommendations for future segmentation quality assessments by analyzing inter-rater agreement between two expert radiologists and correlating their evaluations with established quality metrics, such as the Dice Similarity Coefficient (DSC) and Hausdorff Distance (HD).

Representative examples of expert corrections are shown in Figure~\ref{fig:expert_corrections}, categorized by expert-assigned quality scores of the preliminary automatic segmentations. It is important to note that all cases, including those rated as \textit{Good} during the visual assessment, were sent to the radiologists for manual correction without revealing the quality category (\textit{Good}, \textit{Acceptable}, \textit{Poor}, or \textit{Missed}). This blinded approach ensured that no bias was introduced during the correction process. Green voxels highlight areas added by the experts, while pink voxels indicate areas removed. Minimal corrections were necessary for segmentations rated as \textit{Good}, which typically involved refining tumor boundaries. In contrast, cases categorized as \textit{Missed}, such as \textit{ISPY2\_566011}, required the experts to perform full tumor segmentation from scratch. These examples underscore the critical role of robust automatic segmentation methods in minimizing the burden of manual corrections and highlight the value of expert-corrected segmentations.

\begin{figure}[htb]
    \includegraphics[width=1\linewidth]{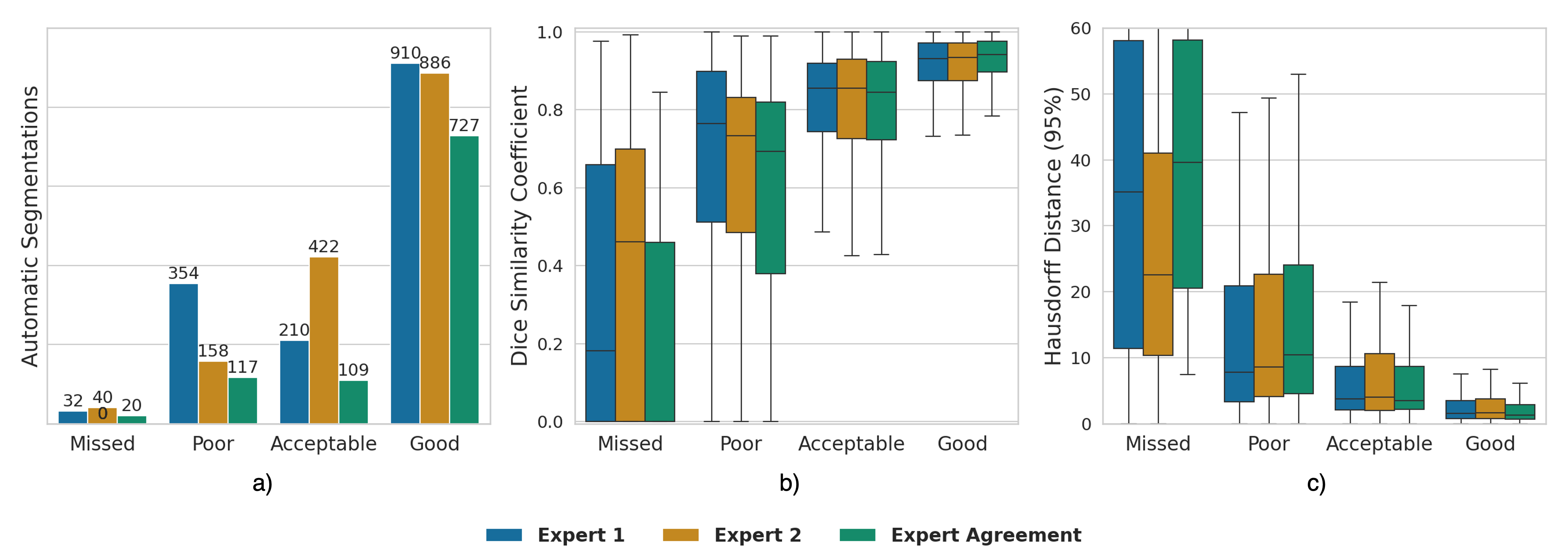}
    \centering
    \caption[]{From left to right: a) distribution of expert-assigned quality scores for automatic segmentations, b) Dice Similarity Coefficient (DSC) and c) the 95 percentile Hausdorff Distance (HD) between automatic and manual segmentations across different quality categories. Panel (b) displays DSC (higher values indicate better agreement), while Panel (c) shows HD (lower values indicate better alignment), respectively.}
    \label{fig:annotation_plots}
\end{figure}

Figure~\ref{fig:annotation_plots} illustrates the distribution of quality scores assigned by the experts, alongside the DSC and HD values calculated between expert-corrected and preliminary automatic segmentations. Segmentations rated as \textit{Good} were strongly correlated with high DSC values and low HD, confirming the reliability of the visual quality control interface in distinguishing segmentations that meet clinical standards. However, the greatest discrepancies between experts occurred in the \textit{Acceptable} and \textit{Poor} categories, where subjective interpretation played a larger role in the quality assessment.

To quantify inter-rater agreement, we computed Cohen’s Kappa under two categorization schemes: the original four categories (\textit{Good}, \textit{Acceptable}, \textit{Poor}, and \textit{Missed}) and a simplified binary scheme (\textit{Good} versus \textit{Corrections Needed}). The four-category scheme yielded a Cohen’s Kappa of 0.39, reflecting low to moderate agreement and indicating inconsistencies, particularly in intermediate categories. Simplifying to a binary scheme improved agreement significantly, with a Cohen’s Kappa of 0.53, demonstrating that a simplified approach improves consistency between experts while maintaining the primary goal of identifying segmentations that require manual correction.

These findings emphasize the efficiency of adopting a binary categorization scheme in visual quality control workflows. By focusing only on segmentations requiring correction (\textit{Corrections Needed}) versus those considered satisfactory (\textit{Good}), future visual assessments can achieve greater inter-rater agreement, reduce annotation time, and optimize the use of expert resources. The results of this analysis aim to contribute to the development of robust visual quality control guidelines for assessing automatic segmentation models.

Furthermore, the categorization of the preliminary automatic segmentations by the radiologists provides a valuable dataset that can be leveraged to train a classifier capable of mimicking radiologists in assessing segmentation quality. Such a classifier could automatically identify \textit{Good} segmentations or flag those requiring corrections, further reducing the reliance on manual evaluations and improving scalability. A comprehensive list of expert quality scores and corresponding distance metrics is included in the Data Records as a CSV file, enabling further analysis, benchmarking efforts, and the potential training of automated quality assessment tools.

\section*{Code Availability}

The GitHub repository for the MAMA-MIA dataset is available at: \href{https://github.com/LidiaGarrucho/MAMA-MIA}{https://github.com/LidiaGarrucho/MAMA-MIA}. This repository provides detailed resources and instructions to support researchers in integrating the MAMA-MIA dataset into their studies. It includes comprehensive scripts to facilitate working with the dataset and enable customization of workflows for segmentation, detection, and analysis in various radiomic and AI-driven studies. The key components of the repository are as follows:

\begin{itemize}
    \item \textbf{Data Filtering and Visualization Scripts:} Jupyter notebooks are provided to guide users in filtering and visualizing clinical and imaging data using the \texttt{pandas} library, along with the clinical and imaging variables included in the dataset's supplementary Excel table. The notebooks include:
    \begin{itemize}
        \item Filtering patients based on clinical outcomes, such as tumor subtype, treatment response, or survival status.
        \item Sorting and grouping cases by scanner manufacturer, image resolution, or pixel spacing.
        \item Exploring relevant clinical and imaging metadata from the dataset information table.
        \item Displaying filtered datasets in tabular or graphical formats for further analysis.
        \item Plotting MRI images using \texttt{matplotlib} for quick visual inspection.
        \item Displaying VOIs and segmentation contours, along with 3D visualizations of tumor segmentations.        
    \end{itemize}
    \item \textbf{Preprocessing Script:} A Jupyter notebook describing common preprocessing steps has been designed to ensure the dataset's compatibility with a wide range of analysis pipelines, while offering flexibility for researchers to adapt workflows to their specific objectives. For a detailed guide, refer to the \href{https://github.com/LidiaGarrucho/MAMA-MIA/tree/main/notebooks}{notebooks directory}. The script includes methods to perform:
    \begin{itemize}
        \item \textbf{Bias Field Correction}: Corrects for intensity inhomogeneities and biases, which can enhance the accuracy of subsequent analyses.
        \item \textbf{Denoising}: A non-linear filter that averages similar patches throughout the image, allowing it to selectively reduce noise without sacrificing edges and subtle intensity variations.
        \item \textbf{Intensity Clipping}: clipping intensity values to a defined percentile range can help reduce outliers.
        \item \textbf{Normalization (Z-score)}: Pixel intensity normalization is performed by subtracting the mean and dividing by the standard deviation, resulting in a distribution with zero mean and unit standard deviation.
        \item \textbf{Histogram Equalization}: Enhances the contrast of the images by redistributing the intensity values to span the full range of the histogram. 
        \item \textbf{Resampling}: e.g. Resampling the images to isotropic voxel size (e.g., $1 \times 1 \times 1$ mm³), ensuring uniform resolution across all images.
    \end{itemize}

    \item \textbf{Bounding Box Extraction:} Code to extract 3D bounding box coordinates containing primary lesions to ease the detection and isolation of VOIs for tumor analysis.
    
    \item \textbf{nnU-Net Segmentation Model Training:} Detailed steps for training an nnU-Net model using the dataset's segmentations. The automatic segmentation models were trained using the \href{https://github.com/MIC-DKFZ/nnU-Net}{nnU-Net} framework in a 5-fold cross-validation setting. Additionally, the pretrained weights of the segmentation model, trained on all expert segmentations, are shared and hosted in the Synapse repository. This facilitates model development and enables easier benchmarking and comparison by other researchers.
\end{itemize}

The original DICOM images from TCIA were transformed to NIfTI format using the \href{https://github.com/amine0110/pycad}{pycad} Python library. Metrics such as the Dice coefficient and 95\% Hausdorff Distance were computed using the \href{https://pypi.org/project/seg-metrics}{seg-metrics 1.2.7} Python library \cite{38}.

\section*{Acknowledgements}
The authors would like to express gratitude to all the participants who contributed to this study. Special appreciation is extended to the experts who performed manual segmentations for 1095 cases and the two experts who conducted quality control on the automatic segmentations. Additionally, our thanks go to Ritse Mann and Marco Caballo for facilitating the 251 manual segmentations from DUKE, as derived from their study~\cite{6}.
We also acknowledge The Cancer Imaging Archive team for making the imaging and clinical data used in this study publicly available, as well as the additional 160 ISPY1 tumor segmentation data from the University of Chicago lab of Maryellen Giger, whose members participated in the TCGA Breast Phenotype Research Group.

\section*{Funding Statement}
This project has received funding from the European Union’s Horizon 2020 research and innovation programmes under grant agreement No 952103 (EUCanImage) and No 101057699 (RadioVal). Also, this work was partially supported by the project FUTURE-ES (PID2021-126724OB-I00) from the Ministry of Science and Innovation of Spain. The co-author K.K. holds the Juan de la Cierva fellowship with a reference number FJC2021-047659-I.

\section*{Author Contributions Statement}
L.G. designed and led the study. C.R. and L.G. collected the dataset, preprocessed the images, curated the clinical and scanner information, verified all the manual segmentations, and contributed to manuscript writing. S.J. and R.O. assisted in curating the manual segmentations from Caballo et al.'s study. S.J. assisted in training the automatic segmentation models and contributed to writing the corresponding section in the paper. R.O. led the section on image synthesis. K.K. designed the evaluation tool used to assess the automatic segmentations. A. T. assisted the experts during the manual segmentation process and built the dataset tables in the paper. J. R. and A.C. provided the expert quality scores of the automatic segmentations in this study. R.M. manually annotated 251 from DUKE dataset. F.P. helped with the data managing in TCIA. M.C., P.M.A., S.W.T., S.S.S., N.O.S., A.M.A, A.K., E.D., G.I., K.N., M.E.K., R.G., M.G., O.L. K.G., M.B. were the sixteen experts that performed the manual segmentations of the missing 1095 cases. K.K., A. T., C. M., M.P.S., K.M., F.S., O.D., L.I., and K.L. supervised the study. K.L. is the PI of the project that funded this research. All authors reviewed the manuscript.

\section*{Competing Interests}
The authors declare no competing interests.

\section*{Tables}
\begin{table}[htb]
\centering
\caption{\label{tab:social_clinical_stats}Social (upper half) and clinical (bottom half) variables of the accumulated dataset, MAMA-MIA, consisting of four breast cancer datasets: ISPY1~\cite{9}, ISPY2~\cite{25}, DUKE~\cite{28}, and NACT~\cite{27}. \emph{Age} is measured in years, \emph{Ethnicity} is categorized in Caucasian/White, African American/Black, Asian and Other (Hispanic, American Indian/Alaskan native, Hawaiian/Pacific Islander, Multiple race) groups, while \emph{BMI} (Body Mass Index) is categorized in the indicated groups using the patient weight and height (if patient height was missing, 1.65cm was used as default). \emph{pCR} stands for pathological Complete Response and N/A for \emph{not available}. 
Last row summarizes the number of cases per dataset and in total.}
\resizebox{\textwidth}{!}{
\begin{tabular}{rlrlrlrlrlrl}
    \toprule
     & & \multicolumn{10}{c}{\textbf{MAMA-MIA}} \\
    \arrayrulecolor{light_grey} \cmidrule(lr){3-12}
    & & \multicolumn{2}{c}{\textbf{ISPY1}} & \multicolumn{2}{c}{\textbf{ISPY2}} & \multicolumn{2}{c}{\textbf{DUKE}} & \multicolumn{2}{c}{\textbf{NACT}} & \multicolumn{2}{c}{\textbf{Total}} \\ 
    \multicolumn{2}{c}{Country} & \multicolumn{2}{c}{United States} & \multicolumn{2}{c}{United States} & \multicolumn{2}{c}{United States} & \multicolumn{2}{c}{United States} & \multicolumn{2}{c}{United States} \\
    \multicolumn{2}{c}{Studies time-period} & \multicolumn{2}{c}{2002 -- 2006} & \multicolumn{2}{c}{2010 -- 2016} & \multicolumn{2}{c}{2000 -- 2014} & \multicolumn{2}{c}{1995 -- 2002} & \multicolumn{2}{c}{1995 -- 2016} \\
    \arrayrulecolor{light_grey} \cmidrule(lr){3-12}
    \multirow{2}{*}{} &  & \# & (\textit{\%})  & \# & (\textit{\%})  & \# & (\textit{\%}) & \# & (\textit{\%}) & \# & (\textit{\%}) \\  
    \arrayrulecolor{light_grey} \toprule
    \multirow{6}{*}{Age} & $<$ 40 & 35 & (\textit{20.5}) & 208 & (\textit{21.2}) & 61 & (\textit{21.0}) & 13 & (\textit{20.3}) & 317 & (\textit{21.0}) \\
      & 40-49 & 62 & (\textit{36.3}) & 300 & (\textit{30.6}) & 104 & (\textit{35.7}) & 25 & (\textit{39.1}) & 491 & (\textit{32.6}) \\
      & 50-59 & 56 & (\textit{32.7}) & 317 & (\textit{32.3}) & 74 & (\textit{25.4}) & 18 & (\textit{28.1}) & 465 & (\textit{30.9}) \\
      & 60-69 & 18 & (\textit{10.5}) & 134 & (\textit{13.7}) & 41 & (\textit{14.1}) & 6 & (\textit{9.4}) & 199 & (\textit{13.2}) \\
      & $>=$ 70 & 0 & (\textit{0.0}) & 18 & (\textit{1.8}) & 11 & (\textit{3.8}) & 2 & (\textit{3.1}) & 31 & (\textit{2.1}) \\
      & N/A & 0 & (\textit{0.0}) & 3 & (\textit{0.3}) & 0 & (\textit{0.0}) & 0 & (\textit{0.0}) & 3 & (\textit{0.2}) \\
    \arrayrulecolor{light_grey} \cmidrule(lr){1-12}
    \multirow{5}{*}{Ethnicity} & Caucasian & 129 & (\textit{75.4}) & 777 & (\textit{79.3}) & 177 & (\textit{60.8}) & 45 & (\textit{70.3}) & 1128 & (\textit{74.9}) \\
      & African American & 31 & (\textit{18.1}) & 116 & (\textit{11.8}) & 91 & (\textit{31.3}) & 3 & (\textit{4.7}) & 241 & (\textit{16.0}) \\
      & Asian & 7 & (\textit{4.1}) & 68 & (\textit{6.9}) & 7 & (\textit{2.4}) & 4 & (\textit{6.2}) & 86 & (\textit{5.7}) \\
      & Other & 2 & (\textit{1.2}) & 16 & (\textit{1.6}) & 14 & (\textit{4.8}) & 3 & (\textit{4.7}) & 35 & (\textit{2.3}) \\
      & N/A & 2 & (\textit{1.2}) & 3 & (\textit{0.3}) & 2 & (\textit{0.7}) & 9 & (\textit{14.1}) & 16 & (\textit{1.1})\\
    \arrayrulecolor{light_grey} \cmidrule(lr){1-12}
    \multirow{7}{*}{BMI} & Underweight & 6 & (\textit{3.5}) & 17 & (\textit{1.7}) & 7 & (\textit{2.4}) & 5 & (\textit{7.8}) & 35 & (\textit{2.3}) \\
      & Normal & 57 & (\textit{33.3}) & 301 & (\textit{30.7}) & 73 & (\textit{25.1}) & 39 & (\textit{60.9}) & 470 & (\textit{31.2}) \\
      & Overweight & 40 & (\textit{23.4}) & 224 & (\textit{22.9}) & 71 & (\textit{24.4}) & 15 & (\textit{23.4}) & 350 & (\textit{23.2}) \\
      & Obesity class I & 23 & (\textit{13.5}) & 155 & (\textit{15.8}) & 50 & (\textit{17.2}) & 5 & (\textit{7.8}) & 233 & (\textit{15.5}) \\
      & Obesity class II & 10 & (\textit{5.8}) & 62 & (\textit{6.3}) & 19 & (\textit{6.5}) & 0 & (\textit{0.0}) & 91 & (\textit{6.0}) \\
      & Obesity class III & 26 & (\textit{15.2}) & 39 & (\textit{4.0}) & 13 & (\textit{4.5}) & 0 & (\textit{0.0}) & 78 & (\textit{5.2}) \\
      & N/A & 9 & (\textit{5.3}) & 182 & (\textit{18.6}) & 58 & (\textit{19.9}) & 0 & (\textit{0.0}) & 249 & (\textit{16.5}) \\
    \arrayrulecolor{light_grey} \cmidrule(lr){1-12}
    \multirow{2}{*}{Implants} & Yes & 1 & (\textit{0.6}) & 29 & (\textit{3.0}) & 0 & (\textit{0.0}) & 0 & (\textit{0.0}) & 30 & (\textit{2.0}) \\
      & No & 170 & (\textit{99.4}) & 951 & (\textit{97.0}) & 291 & (\textit{100.0}) & 64 & (\textit{100.0}) & 1476 & (\textit{98.0}) \\
    \arrayrulecolor{light_grey} \cmidrule(lr){1-12}
    \multirow{2}{*}{\shortstack[r]{Bilateral\\cancer}} & Yes & 3 & (\textit{1.8}) & 20 & (\textit{2.0}) & 7 & (\textit{2.4}) & 0 & (\textit{0.0}) & 30 & (\textit{2.0}) \\
      & No & 168 & (\textit{98.2}) & 960 & (\textit{98.0}) & 284 & (\textit{97.6}) & 64 & (\textit{100}) & 1476 & (\textit{98.0}) \\
    \arrayrulecolor{light_grey} \cmidrule(lr){1-12}
    \multirow{3}{*}{\shortstack[r]{Multifocal\\cancer}} & Yes & 4 & (\textit{2.3}) & 389 & (\textit{39.7}) & 139 & (\textit{47.8}) & 7 & (\textit{10.9}) & 539 & (\textit{35.8}) \\
      & No & 7 & (\textit{4.1}) & 591 & (\textit{60.3}) & 152 & (\textit{52.2}) & 57 & (\textit{89.1}) & 807 & (\textit{53.6}) \\
      & N/A & 160 & (\textit{93.6}) & 0 & (\textit{0.0}) & 0 & (\textit{0.0}) & 0 & (\textit{0.0}) & 160 & (\textit{10.6}) \\
    \arrayrulecolor{light_grey} \cmidrule(lr){1-12}
    \multirow{4}{*}{\shortstack[r]{Tumor\\subtype}} & Luminal & 67 & (\textit{39.2}) & 536 & (\textit{54.7}) & 123 & (\textit{42.3}) & 21 & (\textit{32.8}) & 747 & (\textit{49.6}) \\
      & HER2-enriched & 25 & (\textit{14.6}) & 86 & (\textit{8.8}) & 50 & (\textit{17.2}) & 8 & (\textit{12.5}) & 169 & (\textit{11.2}) \\
      & HER2-pure & 29 & (\textit{17.0}) & 0 & (\textit{0.0}) & 30 & (\textit{10.3}) & 6 & (\textit{9.4}) & 65 & (\textit{4.3}) \\
      & Triple neg. & 45 & (\textit{26.3}) & 358 & (\textit{36.5}) & 85 & (\textit{29.2}) & 11 & (\textit{17.2}) & 499 & (\textit{33.1}) \\
      & N/A & 5 & (\textit{2.9}) & 0 & (\textit{0.0}) & 3 & (\textit{1.0}) & 18 & (\textit{28.1}) & 26 & (\textit{1.8}) \\
    \arrayrulecolor{light_grey} \cmidrule(lr){1-12}
    \multirow{3}{*}{pCR} & Yes & 49 & (\textit{28.7}) & 316 & (\textit{32.2}) & 64 & (\textit{22.0}) & 11 & (\textit{17.2}) & 440 & (\textit{29.2}) \\
      & No & 118 & (\textit{69.0}) & 664 & (\textit{67.8}) & 216 & (\textit{74.2}) & 53 & (\textit{82.8}) & 1051 & (\textit{69.8})\\
      & N/A & 4 & (\textit{2.3}) & 0 & (\textit{0.0}) & 11 & (\textit{3.8}) & 0 & (\textit{0.0}) & 15 & (\textit{1.0}) \\
    \arrayrulecolor{light_grey} \cmidrule(lr){3-4} \cmidrule(lr){5-6} \cmidrule(lr){7-8} \cmidrule(lr){9-10} \cmidrule(lr){11-12} 
      &  & \multicolumn{2}{c}{\textbf{171}} & \multicolumn{2}{c}{\textbf{980}} & \multicolumn{2}{c}{\textbf{291}} & \multicolumn{2}{c}{\textbf{64}} & \multicolumn{2}{c}{\textbf{1506}} \\
    \arrayrulecolor{black} \bottomrule
\end{tabular}
}
\end{table}

\begin{table}[htb]
\centering
\small
\caption{\label{tab:imaging_stats}Image acquisition variables of the accumulated dataset, MAMA-MIA, consisting of four breast cancer datasets: ISPY1~\cite{9}, ISPY2~\cite{25}, DUKE~\cite{28}, and NACT~\cite{27}. The upper half shows general acquistion characteristics, while the bottom half shows specifics of the acquired sequences and slices. \emph{Magnetic field strength} is measured in Tesla (T), while \emph{Slice thickness} and \emph{Pixel spacing} in mm. Other \emph{Scanner models} include MAGNETOM Symphony, SymphonyTim, TrioTim, Verio, Skyra, Sonata, Vision, Vision plus, Prisma fit, Espree from Siemens; Signa Excite, Discovery MR750w, Discovery MR750,  Signa HDx, Optima from GE; and Gyroscan Intera, Ingenia, Achieva, Intera from Philips. Other \emph{Image matrices} or slice sizes (measured in pixels) include [320, 320], [400, 400], [416, 416], [432, 432], [448, 448], [480, 480], [528, 528], [560, 560], [576, 576], [640, 640], and [1024, 1024]. The \emph{Number of phases} include the pre-contrast and all post-contrast phases.
To have a broader overview, the mean value, as well as the minimum and maximum values (in []) are given for a selection of variables. 
Last row summarizes the number of cases per dataset and in total.}
\resizebox{\textwidth}{!}{%
\begin{tabular}{rlrlrlrlrlrl}
    \toprule
     & & \multicolumn{10}{c}{\textbf{MAMA-MIA}} \\
    \arrayrulecolor{light_grey} \cmidrule(lr){3-12}
    & & \multicolumn{2}{c}{\textbf{ISPY1}} & \multicolumn{2}{c}{\textbf{ISPY2}} & \multicolumn{2}{c}{\textbf{DUKE}} & \multicolumn{2}{c}{\textbf{NACT}} & \multicolumn{2}{c}{\textbf{Total}} \\ 
    \multirow{2}{*}{} &  & \# & (\textit{\%})  & \# & (\textit{\%})  & \# & (\textit{\%}) & \# & (\textit{\%}) & \# & (\textit{\%}) \\  
    \arrayrulecolor{light_grey} \toprule
    \multirow{2}{*}{\shortstack[r]{Acquisition\\plane}} & Axial & 0 & (\textit{0.0}) & 980 & (\textit{100.0}) & 291 & (\textit{100.0}) & 0 & (\textit{0.0}) & 1271 & (\textit{84.4}) \\
      & Sagittal & 171 & (\textit{100.0}) & 0 & (\textit{0.0}) & 0 & (\textit{0.0}) & 64 & (\textit{100.0}) & 235 & (\textit{15.6}) \\  
    \arrayrulecolor{light_grey} \cmidrule(lr){1-12}
    \multirow{2}{*}{\shortstack[r]{Magnetic\\field strength}} & 1.5 & 171 & (\textit{100.0}) & 715 & (\textit{73.0}) & 136 & (\textit{46.7}) & 64 & (\textit{100.0}) & 1086 & (\textit{72.1}) \\
      & 3.0 & 0 & (\textit{0.0}) & 265 & (\textit{27.0}) & 155 & (\textit{53.3}) & 0 & (\textit{0.0}) & 420 & (\textit{27.9}) \\
    \arrayrulecolor{light_grey} \cmidrule(lr){1-12}
    \multirow{2}{*}{\shortstack[r]{Fat\\suppression}} & Yes & 170 & (\textit{99.4}) & 976 & (\textit{99.6}) & 290 & (\textit{99.7}) & 64 & (\textit{100.0}) & 1500 & (\textit{99.6}) \\
      & No & 1 & (\textit{0.6}) & 4 & (\textit{0.4}) & 1 & (\textit{0.3}) & 0 & (\textit{0.0}) & 6 & (\textit{0.4}) \\
    \arrayrulecolor{light_grey} \cmidrule(lr){1-12}
    \multirow{3}{*}{\shortstack[r]{Scanner\\manufacturer}} & SIEMENS (S) & 44 & (\textit{25.7}) & 252 & (\textit{25.7}) & 115 & (\textit{39.5}) & 0 & (\textit{0.0}) & 411 & (\textit{27.3}) \\
      & GE & 115 & (\textit{67.3}) & 611 & (\textit{62.3}) & 176 & (\textit{60.5}) & 64 & (\textit{100.0}) & 966 & (\textit{64.1}) \\
      & PHILIPS (P) & 12 & (\textit{7.0}) & 117 & (\textit{11.9}) & 0 & (\textit{0.0}) & 0 & (\textit{0.0}) & 129 & (\textit{8.6}) \\
    \arrayrulecolor{light_grey} \cmidrule(lr){1-12}
    \multirow{4}{*}{\shortstack[r]{Scanner\\model}} & Avanto (S) & 0 & (\textit{0.0}) & 123 & (\textit{12.6}) & 70 & (\textit{24.1}) & 0 & (\textit{0.0}) & 193 & (\textit{12.8}) \\
      & SIGNA HDxt (GE) & 0 & (\textit{0.0}) & 536 & (\textit{54.7}) & 59 & (\textit{20.3}) & 0 & (\textit{0.0}) & 595 & (\textit{39.5}) \\
      & SIGNA GENESIS (GE) & 103 & (\textit{60.2}) & 0 & (\textit{0.0}) & 0 & (\textit{0.0}) & 64 & (\textit{100}) & 167 & (\textit{11.1}) \\
      & Other (S, GE, P) & 68 & (\textit{39.8}) & 321 & (\textit{32.8}) & 162 & (\textit{55.7}) & 0 & (\textit{0.0}) & 551 & (\textit{36.6}) \\
    \arrayrulecolor{light_grey} \cmidrule(lr){1-12}
    \multirow{2}{*}{\shortstack[r]{Bilateral\\MRI}} & Yes & 3 & (\textit{1.8}) & 171 & (\textit{17.4}) & 291 & (\textit{100.0}) & 0 & (\textit{0.0}) & 465 & (\textit{30.9}) \\
      & No & 168 & (\textit{98.2}) & 809 & (\textit{82.6}) & 0 & (\textit{0.0}) & 64 & (\textit{100.0}) & 1041 & (\textit{69.1}) \\
    \arrayrulecolor{light_grey} \cmidrule(lr){1-12}
    \multirow{4}{*}{\shortstack[r]{Image\\matrix}} & [256, 256] & 156 & (\textit{91.2}) & 33 & (\textit{3.4}) & 0 & (\textit{0.0}) & 62 & (\textit{96.9}) & 251 & (\textit{16.7}) \\
      & [384, 384] & 0 & (\textit{0.0}) & 149 & (\textit{15.2}) & 0 & (\textit{0.0}) & 0 & (\textit{0.0}) & 149 & (\textit{9.9}) \\
      & [512, 512] & 15 & (\textit{8.8}) & 721 & (\textit{73.6}) & 176 & (\textit{60.5}) & 2 & (\textit{3.1}) & 914 & (\textit{60.7}) \\
      & Other & 0 & (\textit{0.0}) & 77 & (\textit{7.9}) & 115 & (\textit{39.5}) & 0 & (\textit{0.0}) & 192 & (\textit{12.7}) \\
    \arrayrulecolor{light_grey} \cmidrule(lr){1-12}
    \multirow{3}{*}{\shortstack[r]{Number of\\phases}} & 3 & 146 & (\textit{85.4}) & 0 & (\textit{0}) & 4 & (\textit{1.4}) & 58 & (\textit{90.6}) & 208 & (\textit{13.8}) \\
    & 4-6 & 25 & (\textit{14.6}) & 190 & (\textit{19.4}) & 287 & (\textit{98.6}) & 5 & (\textit{7.8}) & 507 & (\textit{33.7}) \\
      & >= 7 & 0 & (\textit{0.0}) & 790 & (\textit{80.6}) & 0 & (\textit{0.0}) & 1 & (\textit{1.6}) & 791 & (\textit{52.5}) \\
    \arrayrulecolor{light_grey} \cmidrule(lr){2-12}
      & \emph{mean [min, max]} & \textit{3} & \textit{[3, 6]} & \textit{7} & \textit{[4, 11]} & \textit{4} & \textit{[3, 6]} & \textit{3} & \textit{[3, 7]} & \textit{6} & \textit{[3, 11]} \\ 
    \arrayrulecolor{light_grey} \cmidrule(lr){1-12}
    \multirow{3}{*}{\shortstack[r]{Number of\\slices}} & < 100 & 166 & (\textit{97.1}) & 593 & (\textit{60.5}) & 3 & (\textit{1.0}) & 64 & (\textit{100.0}) & 826 & (\textit{54.8}) \\
      & 100-199 & 1 & (\textit{0.6}) & 369 & (\textit{37.7}) & 257 & (\textit{88.3}) & 0 & (\textit{0.0}) & 627 & (\textit{41.6}) \\
      & >= 200 & 4 & (\textit{2.3}) & 18 & (\textit{1.8}) & 31 & (\textit{10.7}) & 0 & (\textit{0.0}) & 53 & (\textit{3.5}) \\
    \arrayrulecolor{light_grey} \cmidrule(lr){2-12}
      & \emph{mean [min, max]} & \textit{64} & \textit{[44, 256]} & \textit{106} & \textit{[52, 256]} & \textit{169} & \textit{[60, 256]} & \textit{60} & \textit{[46, 64]} & \textit{111} & \textit{[44, 256]} \\
    \arrayrulecolor{light_grey} \cmidrule(lr){1-12}
    \multirow{3}{*}{\shortstack[r]{Slice\\thickness}} & < 2.0 & 5  & (\textit{2.9}) & 183 & (\textit{18.7}) & 287 & (\textit{98.6}) & 0 & (\textit{0.0}) & 475 & (\textit{31.5}) \\
      & 2.0-2.9 & 131 & (\textit{76.6}) & 796 & (\textit{81.2}) & 4 & (\textit{1.4}) & 64 & (\textit{100.0}) & 995 & (\textit{66.1})\\
      & >= 3.0 & 35 & (\textit{20.5}) & 1 & (\textit{0.1}) & 0 & (\textit{0.0}) & 0 & (\textit{0.0}) & 36 & (\textit{2.4}) \\
    \arrayrulecolor{light_grey} \cmidrule(lr){2-12}
      & \emph{mean [min, max]} & \textit{2.4} & \textit{[1.5, 4.0]} & \textit{2.0} & \textit{[0.8, 3.0]} & \textit{1.1} & \textit{[1.0, 2.5]} & \textit{2.0} & \textit{[2.0, 2.4]} & \textit{1.9} & \textit{[0.8, 4.0]} \\
    \arrayrulecolor{light_grey} \cmidrule(lr){1-12}
    \multirow{3}{*}{\shortstack[r]{Pixel\\spacing}} & < 0.5 & 15 & (\textit{8.8}) & 2 & (\textit{0.2}) & 0 & (\textit{0.0}) & 2 & (\textit{3.1}) & 19 & (\textit{1.3}) \\
      & 0.5-0.9 & 150 & (\textit{87.7}) & 939 & (\textit{95.8}) & 274 & (\textit{94.2}) & 62 & (\textit{96.9}) & 1425 & (\textit{94.6})\\
      & >= 1.0 & 6 & (\textit{3.5}) & 39 & (\textit{4.0}) &  & (\textit{5.8}) & 0 & (\textit{0.0}) &  & (\textit{4.1}) \\
    \arrayrulecolor{light_grey} \cmidrule(lr){2-12}
      & \emph{mean [min, max]} & \textit{0.8} & \textit{[0.4, 1.2]} & \textit{0.7} & \textit{[0.3, 1.4]} & \textit{0.7} & \textit{[0.5, 1.3]} & \textit{0.7} & \textit{[0.4, 0.9]} & \textit{0.7} & \textit{[0.3, 1.4]} \\
    \arrayrulecolor{light_grey} \cmidrule(lr){3-4} \cmidrule(lr){5-6} \cmidrule(lr){7-8} \cmidrule(lr){9-10} \cmidrule(lr){11-12} 
      &  & \multicolumn{2}{c}{\textbf{171}} & \multicolumn{2}{c}{\textbf{980}} & \multicolumn{2}{c}{\textbf{291}} & \multicolumn{2}{c}{\textbf{64}} & \multicolumn{2}{c}{\textbf{1506}} \\
    \arrayrulecolor{black} \bottomrule
\end{tabular}
}
\end{table}

\begin{table}[htb]
\centering
\footnotesize
\caption{\label{tab:ph_times}Average time intervals, measured in seconds, between the pre-contrast and the subsequent post-contrast phases with the corresponding minimum and maximum values in brackets. Here, details of up to sixth post-contrast phase are shown.}
\resizebox{\textwidth}{!}{
\begin{tabular}{lrlrlrlrlrlrl}
    \toprule
    & \multicolumn{12}{c}{\textbf{Average Time Intervals}} \\
    & \multicolumn{2}{c}{pre to 1\textsuperscript{st} post} & \multicolumn{2}{c}{1\textsuperscript{st} to 2\textsuperscript{nd} post} & \multicolumn{2}{c}{2\textsuperscript{nd} to 3\textsuperscript{rd} post} & \multicolumn{2}{c}{3\textsuperscript{rd} to 4\textsuperscript{th} post} & \multicolumn{2}{c}{4\textsuperscript{th} to 5\textsuperscript{th} post} & \multicolumn{2}{c}{5\textsuperscript{th} to 6\textsuperscript{th} post} \\
    \arrayrulecolor{light_grey} \cmidrule(lr){2-13}
    \textbf{ISPY1} & 390  & [27, 915] & 284 & [20, 531] & 324 & [24, 899] & 142 & [121, 162] & 142 & [121, 162] & -- & -- \\
    \arrayrulecolor{light_grey} \cmidrule(lr){1-13}
    \textbf{ISPY2} & 145  & [77, 761] & 92 & [59, 206] & 92 & [58, 206] & 92 & [58, 217] & 91 & [58, 206] & 90 & [59, 198] \\
    \arrayrulecolor{light_grey} \cmidrule(lr){1-13}
    \textbf{DUKE} & 241 & [94, 922] & 124 & [75, 354] & 114 & [74, 169] & 118 & [88, 391] & -- & -- & -- & --  \\
    \arrayrulecolor{light_grey} \cmidrule(lr){1-13}
    \textbf{NACT} & 442  & [331, 752] & 362 & [290, 901] & 314 & [283, 435] & 288 & [286, 289] & 286 & [286, 286] & 308 & [308, 308] \\
    \arrayrulecolor{black} \cmidrule(lr){1-13}
    \textbf{MAMA-MIA} & 203  & [27, 922] & 131 & [20, 901] & 100 & [24, 899] & 96 & [58, 391] & 91 & [58, 286] & 92 & [59, 308] \\
    \arrayrulecolor{black} \bottomrule
\end{tabular}
}
\end{table}

\end{document}